\documentclass{sig-alternate-05-2015}
\usepackage{times}
\usepackage{helvet}
\usepackage{courier}
\usepackage{booktabs}
\usepackage{epsfig}
\usepackage{amssymb}
\usepackage{amsmath}
\usepackage{amsfonts}
\usepackage{breqn}
\usepackage{multirow}
\usepackage{array}
\usepackage{mathtools}
\usepackage{caption}
\usepackage{url}
\usepackage{float}
\usepackage{balance}
\newcommand{\xhdr}[1]{\vspace{2mm}\noindent{{\bf #1}}}

\newcommand{\eq}[1]{Eq.~\ref{#1}}
\DeclareMathOperator*{\argmax}{arg\,max}

\setcopyright{acmcopyright}
\permission{Copyright is held by the International World Wide Web Conference Committee (IW3C2). IW3C2 reserves the right to provide a hyperlink to the author's site if the Material is used in electronic media.}
\conferenceinfo{WWW 2016,}{April 11--15, 2016, Montr\'eal, Qu\'ebec, Canada.} 
\copyrightetc{ACM \the\acmcopyr}
\crdata{978-1-4503-4143-1/16/04. \\
DOI: http://dx.doi.org/10.1145/2872427.2883037 }

\begin{document}

\title{Ups and Downs: Modeling the Visual Evolution of Fashion Trends with One-Class Collaborative Filtering}

\numberofauthors{2}
\author{
\alignauthor
Ruining He\\
       \affaddr{University of California, San Diego}\\
       \affaddr{La Jolla, California, U.S.A.}\\
       \email{r4he@cs.ucsd.edu}
\alignauthor
Julian McAuley\\
       \affaddr{University of California, San Diego}\\
       \affaddr{La Jolla, California, U.S.A.}\\
       \email{jmcauley@cs.ucsd.edu}
}

\maketitle
\begin{abstract}
Building a successful recommender system depends on understanding both the dimensions of people's preferences as well as their dynamics. In certain domains, such as fashion, modeling such preferences can be incredibly difficult, due to the need to simultaneously model the visual appearance of products as well as their evolution over time. 
The subtle semantics and non-linear dynamics of fashion evolution raise unique challenges especially considering the sparsity and large scale of the underlying datasets.
In this paper we build novel models for the One-Class Collaborative Filtering setting, where our goal is to estimate users' \emph{fashion-aware} personalized ranking functions based on their past feedback. To uncover the complex and evolving visual factors that people consider when evaluating products, our method combines high-level visual features extracted from a deep convolutional neural network, users' past feedback, as well as evolving trends within the community. 
Experimentally we evaluate our method on two large real-world datasets from \emph{Amazon.com}, where we show it to outperform state-of-the-art personalized ranking measures, and also use it to visualize the high-level fashion trends across the 11-year span of our dataset.
\end{abstract}

\frenchspacing
\setlength{\pdfpagewidth}{8.5in}
\setlength{\pdfpageheight}{11in}



\section{Introduction}

\begin{figure}[t]
\centering
\includegraphics[width=\columnwidth]{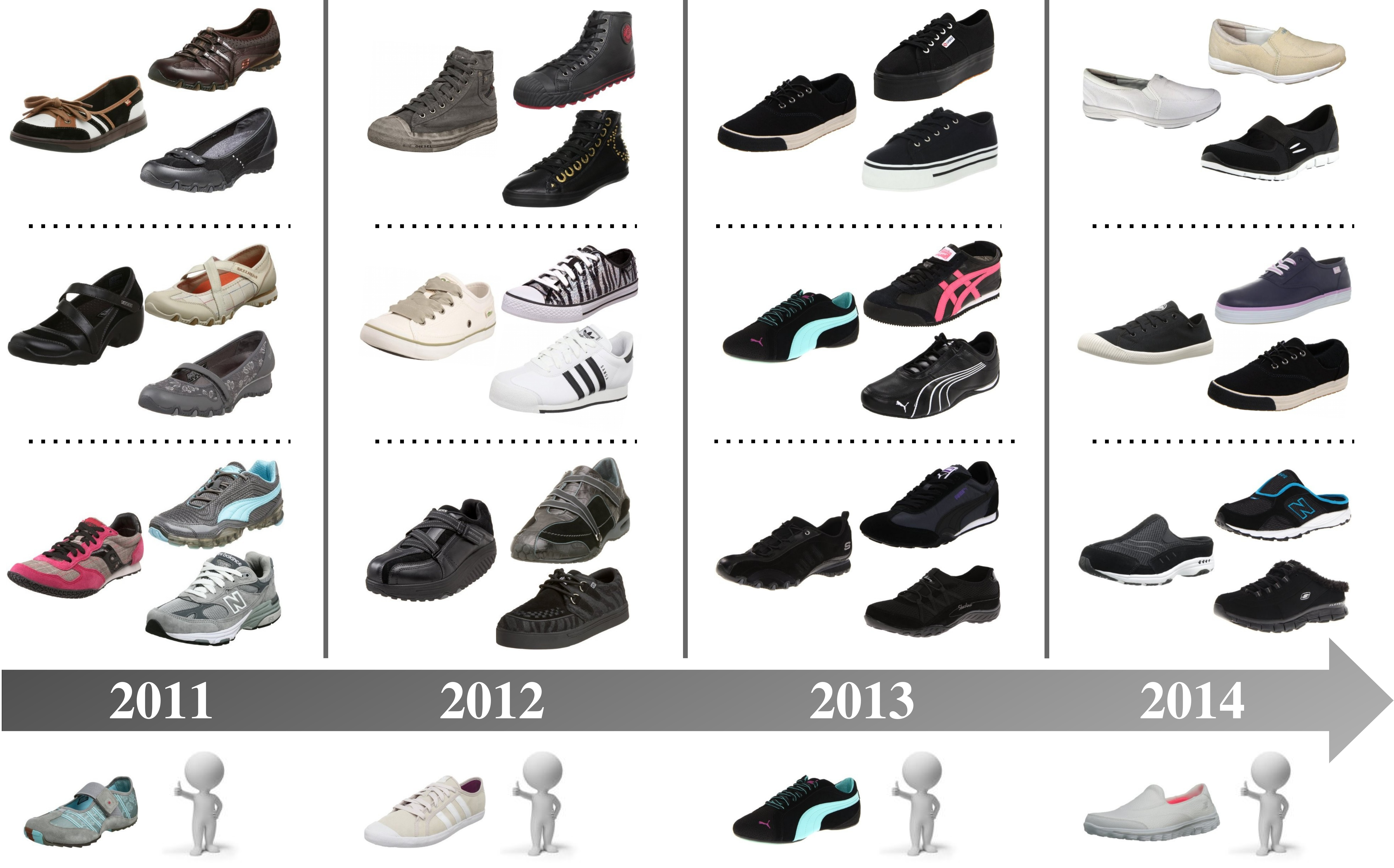}
\caption{Above the timeline are the three most fashionable styles (i.e., groups) of women' sneakers during each year/epoch, revealed by our model; while below the timeline is a specific user's purchases (one in each year), which we model as being the result of a combination of fashion and personal factors.}
\label{fig:timeline}
\end{figure}

Recommender systems play a key role in helping users to discover items matching their personal interests amongst huge corpora of products. In order to surface useful recommendations, it is crucial to be able to learn from user feedback in order to understand and capture the underlying decision factors that have an influence on users' choices. Here we are interested in applications in which \emph{visual} decision factors are at play, such as clothing recommendation. In such settings, visual signals play a key role---naturally one wouldn't buy a t-shirt from \emph{Amazon} without being able to see a picture of the product, no matter what ratings or reviews the product had. Likewise then, when building a recommender system, we argue that this important source of information should be accounted for when modeling users' preferences.

In spite of their potential value, there are several issues that make visual decision factors particularly difficult to model. 
First is simply the complexity and subtlety of the factors involved; to extract any meaningful signal about the role of visual information in users' purchasing decisions shall require large corpora of products (and images) and purchases. Second is the fact that visual preferences are highly personal, so we require a system that models and accounts for the preferences of and differences between individuals. Third is the fact that complex \emph{temporal dynamics} are at play, since the features considered `fashionable' change as time progresses. And finally, it is important to account for the considerable amount of \emph{non}-visual factors that are also at play (such as durability and build quality); this latter point is particularly important when trying to interpret the role of visual decision factors, since we need to `tease apart' the visual from the non-visual components of people's decisions.

Our main goal is to address these four challenges, i.e., to build visually-aware recommender systems that are scalable, personalized, temporally evolving, and interpretable. We see considerable value in solving such problems---in particular we shall be able to build better recommender systems that surface products that more closely match users' and communities' evolving interests. This is especially true for fashion recommendation, where product corpora are particularly `long-tailed' as new items are continually introduced; in such cold-start settings we cannot rely on user feedback but need a rich model of the product's appearance in order to generate useful recommendations.

Beyond generating better recommendations, such a system has the potential to answer high-level questions about how visual features influence people's decisions, and more broadly how fashions have evolved over time. For instance, we can answer queries
such as ``what are the key visual features or factors that people consider when evaluating products?'' or ``what are the main factors differentiating early 2000s vs.~late 2000s fashions?'', or even ``at what point did Hawaiian shirts go out of style?''. Thus our main goal is to learn from data how to model users' preferences toward products, and by doing so to make high-level statements about the temporal and visual dynamics at play.

Addressing our goals above requires new models to be developed. Previous models have considered either visual~\cite{StreetFashion,VBPR} or temporal data~\cite{zhang2015daily,timeSVD,lathia2010temporal,cebrian2010music} in isolation, though few have modeled both aspects simultaneously as we do here.
First, as we show quantitatively, the evolution of fashion trends can be abrupt and non-linear, so that existing temporal models such as timeSVD++~\cite{timeSVD} are not immediately appropriate to address the challenge of capturing fashion dynamics. Moreover, multiple sources of temporal dynamics can be at play simultaneously, e.g.~dynamics at the user or community level; the introduction of new products; or sales promotions that impact the choices people make in the short term.
Thus we need a flexible temporal model that is capable of accounting for these varied effects; this is especially true if we want to interpret our findings, which requires that we `tease-apart' or separate these visual vs.~non-visual
temporal dynamics.
Secondly, real-world datasets are often highly sparse, especially for clothing data where new products are constantly emerging and being replaced over time;
this means on the one hand that accounting for \emph{content} (i.e., visual information) is critical for new items, but on the other hand that only a modest amount of parameters are affordable per item due to the huge item vocabulary involved. This drives us to avoid using localized structures as much as possible.
Thirdly, scalability can be a potential challenge since the new model needs to be built on top of a large corpus of product image data as well as a huge amount of user feedback. Note that the high dimensionality of the image data also exacerbates the above sparsity issue. 

Specifically, our main contributions include:
\begin{enumerate}
\item We build scalable models to capture temporal dynamics in order to make better recommendations for the classical One-Class Collaborative Filtering setting~\cite{OCCF}, where only the implicit (or~`positive') feedback of users (i.e., purchase histories, bookmarks, browsing logs, mouse activities etc.~\cite{dwelltime}) are available. To cope with the non-linearity of fashion trends, we propose to automatically discover the important fashion `epochs' each of which captures a separate set of prevailing visual decision factors at play.
\item Our method also models non-visual dimensions and non-visual temporal dynamics (in a lightweight manner), which not only helps to account for interference from non-visual sources, but also makes our method a fully-fledged recommendation system. 
 We develop efficient training procedures based on the Bayesian Personalized Ranking (BPR) framework to learn the epoch segmentation and model parameters simultaneously. 
\item Empirical results on two large real-world datasets, \emph{Women's} and \emph{Men's Clothing \& Accessories} from \emph{Amazon}, demonstrate that our models are able to outperform state-of-the-art methods significantly, both in warm- and cold-start settings. 
\item We provide visualizations of our learned models and qualitatively demonstrate how fashion has shifted in recent years. We find that fashions evolve in complex, non-linear ways, which can not easily be captured by existing methods.
\end{enumerate}

The rest of the paper is organized as follows. We introduce our proposed method in Section~\ref{sec:model}, before we develop a Coordinate Ascent fitting procedure in Section~\ref{sec:learn}. Comprehensive experiments on real-world datasets as well as visualizations are conducted in Section~\ref{sec:experiment}. We discuss related work in Section~\ref{sec:relwork} and conclude in Section~\ref{sec:conclusion}.

\section{Modeling the Temporal Dynamics of Visual Styles} \label{sec:model}
We are interested in learning visual temporal dynamics from implicit feedback datasets (e.g.~purchase histories of clothing \& accessories) where visual signals are at play, rather than (say) star-ratings. This choice is made due to the expectation that evolving fashion styles will be more closely reflected in purchase choices than in ratings---our hypothesis being that people only buy items if they are already attracted to their visual appearance, so that variation in ratings can be predominantly explained by \emph{non}-visual factors, whereas variation in purchases is a combination of both visual and non-visual decisions.

By accounting for evolving fashion dynamics for implicit feedback in the form of purchase histories, we hope to build systems that are quantitatively helpful for estimating users' personalized rankings (i.e., assigning likely purchases higher ranks than non-purchases), which can then be harnessed for recommendation.

\begin{table}
\caption{Notation \label{tab:notation}}
\begin{tabular}{lp{0.7\linewidth}} \toprule
Notation & Explanation\\ \midrule
$\mathcal{U}$, $\mathcal{I}$ & user set, item set\\ 
$\mathcal{I}_u^+$ & the items for which user $u$ expressed positive feedback\\
$\mathcal{P}_u$,$\mathcal{V}_u$, $\mathcal{T}_u$ & training/validation/test subsets of $\mathcal{I}_u^+$\\
$\widehat{x}_{u,i}$ & predicted preference of user $u$ towards item $i$\\
$\widehat{x}_{u,i}(t)$ & predicted preference of $u$ towards $i$ at time $t$\\
$K$ & dimensionality of latent factors\\
$K'$ & dimensionality of visual factors\\
$F$ & dimensionality of Deep CNN features\\
$\alpha$ & global offset (scalar)\\
$\beta_u$, $\beta_i$ & user $u$'s bias, item $i$'s bias (scalar)\\
$\beta_i(t)$ & item $i$'s bias at time $t$ (scalar)\\
$\beta_{C_i}(t)$ & subcategory bias item at time $t$ (scalar)\\
$\gamma_u$, $\gamma_i$ & latent factors of user $u$, item $i$ ($K \times 1$)\\
$\theta_u$, $\theta_i$ & visual factors of user $u$, item $i$ ($K' \times 1$)\\
$\theta_u(t)$, $\theta_i(t)$ & visual factors of user $u$, item $i$ at time $t$ ($K' \times 1$)\\
$f_i$ & Deep CNN visual features of item $i$ ($F \times 1$)\\
$\mathbf{E}$ & $K' \times F$ embedding matrix\\
$\mathbf{E}(t)$ & $K' \times F$ embedding matrix at time $t$\\
$\beta$ & visual bias vector (visual bias = $\langle \beta, f_i \rangle $)\\
$\beta(t)$ & visual bias vector at $t$ (visual bias = $\langle \beta(t), f_i \rangle $)\\
\bottomrule
\end{tabular}
\end{table}

Formally, we represent the set of users and items with $\mathcal{U}$ and $\mathcal{I}$ respectively. Each user $u \in \mathcal{U}$ is associated with a set of items $\mathcal{I}_u^+$. About each item $i \in \mathcal{I}_u^+$, $u$ has expressed explicit positive feedback (i.e., by purchasing it) at time $t_{ui}$. Additionally, a single image is available for each item $i \in \mathcal{I}$. Using the above data, our objective is to generate for each user $u$ a \emph{time-dependent} personalized ranking of those items about which they haven't yet provided feedback (i.e. $\mathcal{I} \setminus \mathcal{I}_u^+$). The challenge here is to develop efficient methods to make use of these raw images to learn visual styles that are temporally-evolving and predictive of users' opinions. The notation we use throughout the paper is summarized in Table \ref{tab:notation}.

\subsection{Matrix Factorization}
We begin by briefly describing the underlying `standard' Matrix Factorization method~\cite{korenSurvey}, whose basic formulation we adopt. Here the preference of a user $u$ toward an item $i$ (i.e. $\widehat{x}_{u,i}$) is predicted according to
\begin{equation} \label{eq:baseline}
\widehat{x}_{u,i} = \alpha + \beta_u + \beta_i + \langle \gamma_u, \gamma_i \rangle,
\end{equation}
where $\alpha$ is a global offset, $\beta_u$ and $\beta_i$ are user/item bias terms, and $\gamma_u$ and $\gamma_i$ are $K$-dimensional latent factors describing user $u$ and item $i$ respectively. Intuitively, $\gamma_i$ can be interpreted as the `properties' of the item $i$, while $\gamma_u$ can be seen as user $u$'s personal `preferences' toward those properties.

\subsection{Modeling Visual Dimensions} \label{subsec:VBPR}
Although the above standard model can capture rich interactions between users and items, it suffers from \emph{cold start} issues due to the sparsity of real-world datasets, especially in domains like fashion where the product vocabulary is long-tailed and continuously evolving. Using explicit features like user profiles and product features can alleviate this problem by making use of auxiliary signals in cold start scenarios. 

To model visual dimension and uncover users' preferences towards different visual styles, we are interested in incorporating the \emph{visual appearance} of items into the formulation.
Previous methods for `visually aware' recommendation have made use of features from deep networks
\cite{VisualSIGIR,VBPR} though made no use of temporal dynamics. In those works the basic idea is to discover low-dimensional `visual decision factors' to explain user's activities. We build upon this idea and define our predictor as
\begin{equation} \label{eq:VBPR}
\widehat{x}_{u,i} = \underbrace{\alpha + \beta_u + \beta_i}_{\text{bias terms}} + \underbrace{\langle \gamma_u, \gamma_i \rangle}_{\text{non-visual interaction}} + \underbrace{\langle \theta_u, \theta_i \rangle}_{\text{visual interaction}},
\end{equation}
where 
$\alpha$, $\beta$, and $\gamma$ are as in \eq{eq:baseline}. $\theta_u$ and $\theta_i$ are newly introduced $K'$-dimensional \emph{visual factors} that encode the `visual compatibility' between the user $u$ and the item $i$.

Intuitively, we want $\theta_i$ to be explicit visual features of the item $i$. Particularly, it is more desirable to use high-level features to capture human notions of visual styles. Deep Convolutional Neural Network (i.e., `Deep CNN') features extracted from raw product images presented a good option due to their widely demonstrated efficacy at capturing abstract notions of fine-grained categories~\cite{ImageNetchallenge}, photographic style~\cite{ImageStyle}, aesthetic quality~\cite{AestheticsDL}, and scene characteristics~\cite{Decaf}, among others. 

Let $f_i$ denote the Deep CNN features of item $i$ and $F$ represent its number of dimensions. We further introduce a $K' \times F$ embedding matrix $\mathbf{E}$ to linearly embed the high-dimensional feature vector $f_i$ into a much lower-dimensional (i.e., $K'$) \emph{visual style space}. Namely, we take
\begin{equation}
\theta_i = \mathbf{E} f_i.
\end{equation}
Then the parameter set is $\Theta = \{\alpha, \beta_u, \beta_i, \gamma_u, \gamma_i, \theta_u, \mathbf{E}\}$. By learning the embedding $\mathbf{E}$ from the data, we are uncovering $K'$ visual dimensions that are the most predictive of users' opinions. 

\subsection{Modeling Visual Evolution}
The above model is good at capturing/uncovering visual dimensions as well as the extent to which users are attracted to each of them. Nevertheless, fashions, i.e., the visual elements of items that people are attracted to, evolve gradually over time. This presents challenges when modeling the visual dimensions of opinions because the same appearance may be favored during some time periods while disliked during others.
Our goal here is to discover such trends both as a means of making better predictions, but also so that we can draw high-level conclusions about how fashions have evolved over the life of our dataset.

Thus we want to extend the above `static' model to capture the temporal dynamics of fashion. Considering the sparsity of real-world datasets, it is important to develop models that are expressive enough to capture the relevant dynamics but at the same time are tractable in terms of the number of parameters involved.

\subsubsection{Temporally-evolving Visual Factors} \label{subsec:tempfactor}
Here we identify three main fashion dynamics from which we can potentially benefit. We propose models to capture each of them with temporally-evolving visual factors; 
that is we model user/item visual factors as a function of time $t$, i.e.,~$\theta_u(t)$ and $\theta_i(t)$, with their inner products accounting for the temporal user-item visual interactions. This formulation is able to capture different kinds of fashion dynamics as described below.

\xhdr{Temporal Attractiveness Drift.}
The first notion of temporal dynamics is based on the observation that items gradually gain/lose `attractiveness' in different visual dimensions as time goes by. To capture such a phenomenon, it is natural to extend our embedding matrix $\mathbf{E}$ to be time-dependent. More specifically, we model our embedding matrix at time $t$ as 
\begin{equation}
\mathbf{E}(t) = \mathbf{E} + \mathbf{\Delta_E}(t).
\end{equation}
Here the underlying `stationary' component of the model is captured by $\mathbf{E}$ while the time-dependent `drifting' component is accounted for by $\mathbf{\Delta_E}(t)$. Then item $i$'s visual factors at time $t$ become
\begin{equation}
\theta_i(t) = \mathbf{E}(t) f_i.
\end{equation}
In this way, we are modeling fashion evolution across entire communities with \emph{global} low-rank structures. Such structures are expressive while introducing only a modest number of parameters.

\xhdr{Temporal Weighting Drift.}
As fashion evolves over time, it is likely that users \emph{weigh} visual dimensions differently.
For example, people may pay less attention to a dimension describing colorfulness as communities become more tolerant of bright colors. 
Accordingly, we introduce a $K'$-dimensional temporal weighting vector $w(t)$ to capture users' evolving emphasis on different visual dimensions, namely
\begin{equation} \label{eq:vis}
\theta_i(t) = \mathbf{E} f_i \odot w(t),
\end{equation}
where $\odot$ is the Hadamard product. 

Combining the above two dynamics, our formulation for item visual factors becomes
\begin{equation} \label{eq:thetait}
\theta_i(t) = \underbrace{\mathbf{E} f_i \odot  w(t)}_{\text{base}} + \underbrace{\mathbf{\Delta_E}(t) f_i}_{\text{deviation}} 
\end{equation}
such that (when properly regularized) temporal variances are partly explained by the weighting scheme while the rest are absorbed by the expressive deviation term.

Note that compared to our basic model, so far we have only introduced \emph{global} structures that are shared by all users. This achieves our goal of capturing temporal fashion trends that apply to the entire population. Next, we introduce `local' dynamics, in order to model the drift of \emph{personal} tastes over time. 

\begin{figure*}[!t]
\begin{equation} \label{eq:final}
\underbrace{\widehat{x}_{u,i}(t)}_{\substack{\text{preference of user $u$}\\\text{towards item $i$ at time $t$}}} = \underbrace{\alpha + \beta_u + \underbrace{\beta_i(t) + \beta_{C_i}(t)}_{\text{temporal non-visual biases}} + \underbrace{\langle \overbrace{\beta(t)}^{\text{defined by \eq{eq:betat}}}, f_i \rangle}_{\text{temporal visual bias}}}_{\text{bias terms}} + 
\underbrace{\underbrace{\langle \gamma_u, \gamma_i \rangle}_{\text{non-visual interaction}} + \underbrace{\langle \overbrace{\theta_u(t)}^{\text{defined by \eq{eq:thetaut}}}, \overbrace{\theta_i(t)}^{\text{defined by \eq{eq:thetait}}} \rangle}_{\text{temporal visual interaction}}}_{\text{user-item interactions}}.
\end{equation}
\caption{The proposed fashion-aware preference predictor.}
\end{figure*}

\xhdr{Temporal Personal Drift.}
Apart from the above global temporal dynamics (i.e., fashion evolution), there also exist dynamics at the level of drifts in personal tastes over time. In other words, users' opinions are affected by `outside' fashion trends as well as their own personal preferences, both of which can evolve gradually. Modeling this kind of drift can borrow ideas from existing works (e.g.~timeSVD++~\cite{timeSVD}) in order to extend our basic model with time-evolving user visual factors, i.e., by modeling $\theta_u$ as a function of time.
Here we give one example formulation (see~\cite{timeSVD} for more details) as follows:
\begin{equation} \label{eq:thetaut}
\theta_u(t) = \theta_u + sign(t-t_u) \cdot |t-t_u|^{\kappa} \eta_u ,
\end{equation}
which uses a simple parametric form to account for the deviation of user $u$ at time $t$ from his/her mean feedback date $t_u$. This method uses two vectors $\theta_u$ and $\eta_u$ to model each user, with hyperparameter $\kappa$ learned with a validation set (to be described later).

\subsubsection{Temporally-evolving Visual Bias} \label{subsec:tempbias}
In addition to temporally evolving factors $\theta_i(t)$, we introduce a temporal visual bias term to account for that portion of the variance which is common to all factors. 
More precisely, we use a time-dependent $F$-dimensional vector $\beta(t)$ that adopts a formulation resembling that of~\eq{eq:thetait}:
\begin{equation} \label{eq:betat}
\beta{(t)} = \beta \odot  b(t) + \Delta_{\beta}(t).
\end{equation} 
Then the visual bias of item $i$ at time $t$ is computed by taking the inner product $\langle \beta(t), f_i \rangle$.
The intention is to use low-rank structures to capture the changing `overall' response to the appearance, so that the rest of the variance (i.e., per-user and per-dimension dynamics) are captured by properly regularized higher-rank structures, namely the inner product of $\theta_u(t)$ and $\theta_i(t)$. 
Experimentally, incorporating this term improves the performance to some degree, and is also useful for visualization.

\subsubsection{Non-Visual Temporal Dynamics} \label{subsec:nonfashion}
Up to now, we have described how to extend our basic formulation to model visual dynamics. However, there also exist non-visual temporal dynamics in the datasets, such as sales, promotions,
or the emergence of new products. Incorporating such dynamics into our model can not only improve predictive performance, but also helps with interpretability by allowing us to tease apart visual from non-visual decision factors.
Here we want to distinguish as much as possible those factors that can be determined by the item's non-visual properties (such as its category) versus those that can only be determined from the image itself. 

To serve this purpose, we propose to incorporate the following two non-fashion dynamics in a lightweight manner, i.e., we guarantee that we are only introducing an affordable amount of additional parameters due to the sparsity of the real-world datasets we consider. 

\xhdr{Per-Item Temporal Dynamics.}
The first dynamics to model are on the per-item level. As said before, various factors can cause an item to be purchased during some periods and not during others. Our choice is to replace the stationary item bias term $\beta_i$ in~\eq{eq:thetait} with a temporal counterpart $\beta_i(t)$~\cite{timeSVD}.

\xhdr{Per-Subcategory Temporal Dynamics.}
Next, for datasets where the category tree is available (as is the case for the ones we consider), it is also possible to incorporate per-subcategory temporal dynamics.
By accounting for category information explicitly as we do here, we discourage the visual component of our model from indirectly trying to predict the subcategory of the product, so that it may instead focus on subtler visual aspects.
Letting $C_i$ denote the subcategory the item $i$ belongs to, we add a temporal subcategory bias term $\beta_{C_i}(t)$ to our formulation to account for the drifting of users' opinions towards a subcategory.

Gluing all above components together, we predict $\widehat{x}_{u,i}(t)$, the affinity score of user $u$ and item $i$ at time $t$, with \eq{eq:final}.\footnote{Note that when computing personalized rankings for a single user $u$, $\alpha$ and $\beta_u$ in \eq{eq:final} can be ignored.}
Experimentally, we found that \emph{global} temporal dynamics (i.e., fashion trends) are particularly useful at addressing personalized ranking tasks. However, modeling user terms, i.e., temporal personal drift, had relatively little effect in our datasets. The reasons are dataset-specific: (a) our datasets span a decade and most users only remain active during a relatively short period of time; (b) our datasets are highly sparse which means that the lack of per-user observations makes it difficult to fit the high-dimensional models required (see~\eq{eq:thetaut}). 
Therefore for our experiments we ultimately adopted stationary user visual factors $\theta_u$ (note this way users' preferences are still affected by fashion trends). %

\subsubsection{Fashion Epoch Segmentation}

So far we have described \emph{what} temporal components to use in the formulation of our time-aware predictor; what remains to be seen is how to model the temporal term, i.e., how $\beta(t), \theta(t)$ change as time progresses.
One solution is to adopt a fixed schedule to describe the underlying evolution, e.g.~to fit some parameterized function of (say) the raw timestamp, as is done by timeSVD++~\cite{timeSVD}. However, fashion tends to evolve in a non-linear and somewhat abrupt manner, which goes beyond the expressive power of such methods (we experimentally tried parameterized functions like those in timeSVD++ but without success). Instead, a time-window design which uncovers fashion `stages' or `epochs' during the life span of the dataset proved preferable in our case. In other words, we want to \emph{learn} a temporal partition of the timeline of our data into discrete segments during which different visual characteristics predominate to influence users' opinions.

To achieve our goal, we learn a partition of the timeline of our dataset, consisting of $N$ epochs, and to each epoch $\mathit{ep}$ we attach a set of parameters 
$$\Theta_{ep} = \{\mathbf{\Delta_E}(ep), \Delta_{\beta}(ep), w(ep), b(ep), \beta_i(ep), \\ \beta_{C_i}(ep) \}.\footnote{i.e., discretized $\mathbf{\Delta_E}(t), \Delta_{\beta}(t), w(t), b(t), \beta_i(t), \beta_{C_i}(t)$ (respectively).}$$ 
Then we predict the preference of user $u$ towards item $i$ at epoch $ep$ according to
$\widehat{x}_{u,i}(ep(t))$, where the function $ep(\cdot)$ returns the epoch index of time $t$ according to the segmentation. 
Note that while such a model could potentially capture seasonal effects (given fine-grained enough epochs), this is not our goal in this paper since we want to uncover long-term temporal drift; this can easily be achieved by tuning the number of epochs such that they tend to span multiple seasons (e.g.~we obtained the best performance using 10 epochs in our 11 year dataset).

Finally, there are two components of the model to be estimated: (a) the model parameters $\Theta = \cup_{ep}{\Theta_{ep}} \cup \{\alpha, \beta_u, \gamma_u, \gamma_i, \theta_u, \mathbf{E}, \beta \}$, and (b) the fashion epochs themselves, i.e., a partition $\Lambda$ of the timeline into segments with different visual rating behavior.

\section{Learning the Model} \label{sec:learn}
With the above temporal preference predictor, our objective is for each user $u$ to generate a personalized ranking of the items they haven't interacted with (i.e., $\mathcal{I} \setminus \mathcal{I}_u^+$) at time $t$.
Here we adopt Bayesian Personalized Ranking, a state-of-the-art ranking optimization framework~\cite{BPR}, to directly optimize the rankings produced by our model. First we derive the likelihood function we are trying to maximize according to BPR, before we describe the coordinate ascent optimization procedure to learn the fashion epoch segmentation as well as the model parameters.

\subsection{Log-Likelihood Maximization}
Bayesian Personalized Ranking (BPR) is a pairwise ranking optimization framework which adopts Stochastic Gradient Ascent to optimize the regularized corpus likelihood~\cite{BPR}. Let $\mathcal{P}_u \subset \mathcal{I}_u^+$ be the set of positive (i.e., observed) items for user $u$ in the training set. Then according to BPR, a training tuple set $D_S$ consists of triples of the form $(u,i,j)$, where $i \in \mathcal{P}_u$ and $j \in \mathcal{I} \setminus \mathcal{P}_u$. 
Given a triple $(u,i,j) \in D_S$, BPR models the probability that user $u$ prefers item $i$ to item $j$ with $\sigma (\widehat{x}_{u,i} - \widehat{x}_{u,j})$, where $\sigma$ is the sigmoid function, and learns the parameters by maximizing the regularized log-likelihood function as follows:
$$
\sum_{\mathclap{(u,i,j) \in D_S}} \quad \log \sigma (\widehat{x}_{u,i} - \widehat{x}_{u,j}) - \frac{\lambda_{\Theta}}{2} ||\Theta||^2.
$$

Building on the above formulation, we want to add a temporal term
$t_{ui}$ encoding the time at which user $u$ expressed positive feedback about $i \in \mathcal{P}_u$. The basic idea is that we want to rank the observed item $i$ higher than all non-observed items at time $t_{ui}$. More precisely, our training set $D_{S^+}$ is comprised of quadruples of the form $(u,i,j, t_{ui})$, where user $u$ expressed positive feedback about item $i$ at time $t_{ui}$ with $j$ being a non-observed item:
\begin{equation} \label{eq:DS}
D_{S^+} = \{(u,i,j,t_{ui})|u \in \mathcal{U} \wedge i \in \mathcal{P}_u \wedge j \in \mathcal{I} \setminus \mathcal{P}_u \}.
\end{equation}

To simplify this notion, we introduce the shorthand
$$\widehat{x}_{uij}(ep(t_{ui})) = \widehat{x}_{u,i}(ep(t_{ui}))-\widehat{x}_{u,j}(ep(t_{ui})),$$ where $ep(t)$ returns the index of the epoch that timestamp $t$ falls into, and $\widehat{x}_{u,i}(ep)$ as well as $\widehat{x}_{u,j}(ep)$ are defined by~\eq{eq:final}. 
Then according to the BPR framework, our model is fitted by maximizing the regularized log-likelihood of the corpus (i.e., BPR-OPT in~\cite{BPR}): 
\begin{equation} \label{eq:obj}
\widehat{\Theta}, \widehat{\Lambda} = 
\argmax_{\Theta, \Lambda} \quad \sum_{\mathclap{\qquad(u,i,j, t_{ui}) \in D_{S^+}}} \quad \log \sigma (\widehat{x}_{uij}(ep(t_{ui}))) - \frac{\lambda_{\Theta}}{2} ||\Theta||^2.
\end{equation}
Again, note that there are two components to fit to maximize the above objective function, with one being the parameter set $\Theta$ and the other being the segmentation $\Lambda$ of the timeline comprising $N$ fashion epochs. Next we describe how to derive a coordinate-ascent-style optimization procedure to fit these two components.

\subsection{Coordinate Ascent Fitting Procedure}
We adopt an iterative optimization procedure which alternates between (a) fitting the model parameters $\Theta$ (given the segmented timeline $\Lambda$), and (b) segmenting the timeline $\Lambda$ (given the current estimate of the model parameters $\Theta$). This procedure resembles the one used in~\cite{WWWUserExper}, though the problem setting and data are different.

\subsubsection{Fitting the Model Parameters $\Theta$}
This step fixes the epoch segmentation $\Lambda$ and adopts stochastic gradient ascent to optimize the regularized log-likelihood in~\eq{eq:obj}. Given a randomly sampled training quadruple $(u,i,j,t_{ui}) \in D_{S^+}$, the update rule of $\Theta$ is derived as 
\begin{equation}
\Theta \leftarrow \Theta + \epsilon \cdot (\sigma(-\widehat{x}_{uij}(ep(t_{ui}))) \frac{\partial \widehat{x}_{uij}(ep(t_{ui}))}{\partial \Theta} - \lambda_{\Theta}\Theta ), 
\end{equation}
where $\epsilon$ is the learning rate. Sampling strategies may affect the performance of the model to some extent. In our implementation, we sample users uniformly to optimize the average AUC metric (to be discussed later).

\subsubsection{Fitting the Fashion Epoch Segmentation $\Lambda$}
Given the model parameters $\Theta$, this step finds the optimal segmentation of the timeline to optimize the objective in~\eq{eq:obj}.
To achieve this goal, we first partition the timeline into $N$ continuous bins of equal size. Then the fitting problem is solved with a dynamic programming procedure, which finds the segmentation such that rankings inside all bins are predicted most accurately. This is a canonical instance of a sequence segmentation problem~\cite{dynamic}, which admits an
$\mathcal{O}(|\mathcal{D_S^+}| \times N)$ solution in our case. 

\xhdr{Scaling to large datasets.}
Fitting the epoch segmentation in a na\"{\i}ve way would be time-consuming due to
the fact that the `ranking quality' has to be evaluated by enumerating \emph{all} non-observed items for each positive item.
Fortunately, it turns out that for this step we can \emph{approximate} the full log-likelihood by sampling a relatively small `batch' of non-observed items for each positive user-item pair. Experimentally this proved to be effective and allows the dynamic programming procedure to find the optimal solution within around 3 minutes on our largest datasets.

Finally, our parameters are randomly initialized between 0 and 1.0. 
The two fitting steps above are repeated until convergence, or until no further improvement is obtained on the validation set.
We discuss scalability further in Appendix \ref{appd:scale}.

\section{Experiments} \label{sec:experiment}
We perform experiments on two real-world datasets to investigate the efficacy
of our proposed method. First we introduce the datasets we work with, before we compare and evaluate our method against different baselines, and finally visualize the fashion dynamics captured by our model.  

\subsection{Datasets}
To evaluate the strength of our method at capturing fashion dynamics, we are interested in real-world datasets that (a) are broad enough to capture the general tastes of the public, and (b) temporally span a long period so that there are discernibly different visual decision factors at play during different times.

The two datasets we use are from \emph{Amazon.com}, as introduced in~\cite{VisualSIGIR}. We consider two large categories that naturally encode fashion dynamics (within the U.S.) over the past decade, namely Women's and Men's Clothing \& Accessories, each consisting of a comprehensive vocabulary of clothing items. The images available from this dataset are of high quality (typically centered on a white background) and have previously been shown to be effective for recommendation tasks (though different from the one we consider here).

We process each dataset by taking users' review histories as implicit feedback and extracting visual features $f_i$ from one image of each item $i$. We discard users $u$ who have performed fewer than 5 actions, i.e., for whom $|\mathcal{I}_u^+| < 5$. 
Statistics of our datasets are shown in Table~\ref{table:dataset}.

\begin{table}
\centering
\renewcommand{\tabcolsep}{3.8pt}
\caption{Dataset statistics (after processing)} 
\begin{tabular}{lrrrc} \toprule
Dataset         &\#users    &\#items     &\#feedback   & Timespan \\ \midrule
\emph{Women}    & 99,748	& 331,173    & 854,211	   & Mar. 2003 - Jul. 2014\\
\emph{Men}      & 34,212    & 100,654    & 260,352     & Mar. 2003 - Jul. 2014\\ \midrule
Total           & 133,960   & 431,827    & 1,114,563   & Mar. 2003 - Jul. 2014\\ \bottomrule
\end{tabular}
\label{table:dataset}
\end{table}

\subsection{Visual Features}
To extract a visual feature vector $f_i$ for each item $i$ in the above datasets, we employ a pre-trained convolutional neural network, namely the Caffe reference model~\cite{Caffe}, which has previously been demonstrated to be useful at capturing the properties of images of this type \cite{VisualSIGIR}. This model implements the architecture proposed by~\cite{DeepCNNArchitecture} with 5 convolutional layers followed by 3 fully-connected layers and was pre-trained on 1.2 million ImageNet (ILSVRC2010) images. We obtain our $F = 4096$ dimensional visual features by taking the output of the second fully-connected layer (i.e., FC7).

\subsection{Evaluation Methodology}
Given a user-item pair $(u,i)$, the preference of $u$ towards $i$ is a function of time, i.e., the recommended item ranking for $u$ is time-dependent. Therefore for a held-out triple $(u,i,t_{ui})$, our evaluation consists of calculating how accurately item $i$ is ranked for user $u$ at time $t_{ui}$. 

Each of our datasets is split into training/validation/test sets by uniformly sampling for each user $u$ from $\mathcal{I}_u^+$ an item $i$ (associated with a timestamp $t_{ui}$) to be used for validation $\mathcal{V}_u$ and another for testing $\mathcal{T}_u$. The rest of the data $\mathcal{P}_u$ is used for training, 
i.e., $\mathcal{I}_u^+=\mathcal{P}_u \cup \mathcal{V}_u \cup \mathcal{T}_u$ and $|\mathcal{P}_u| = |\mathcal{V}_u| = |\mathcal{U}|$. 

All methods are then evaluated on $\mathcal{T}_u$ with the widely used AUC (\emph{Area Under the ROC curve}) measure:
\begin{equation}
\mathit{AUC} =  \frac{1}{|\mathcal{U}|}  \sum_u   \frac{1}{|E(u)|}   \sum_{(i,j) \in E(u)}  \delta (\widehat{x}_{u,i}(t_{ui}) > \widehat{x}_{u,j}(t_{ui})),
\end{equation}
where the indicator function $\delta(b)$ returns 1 $\mathit{iff}$ $b$ is $\mathit{true}$, and the evaluation goes through the pair set of each user $u$:
\begin{equation}
E(u) = \{(i,j) | i \in \mathcal{T}_u \wedge j \notin  (\mathcal{P}_u \cup \mathcal{V}_u \cup \mathcal{T}_u) \}.
\end{equation}

For all methods we select the best hyperparameters using the validation set $\mathcal{V} = \cup_{u \in \mathcal{U}} \mathcal{V}_u$ and report the corresponding performance on the test set $\mathcal{T} = \cup_{u \in \mathcal{U}} \mathcal{T}_u$.

\subsection{Comparison Methods}
Matrix Factorization (MF) based methods are currently state-of-the-art for modeling implicit feedback datasets (e.g.~\cite{BPR,GBPR,MRBPR}). Therefore we mainly compare against state-of-the-art MF methods in this area, including both point-wise and pairwise MF models (see Section \ref{sec:relwork} for more details).
\begin{itemize} 
\item \textbf{Popularity (POP):} Items are ranked according to their popularity.
\item \textbf{WR-MF:} A state-of-the-art point-wise MF model for implicit feedback proposed by~\cite{WRMF}.
It assigns confidence levels to different feedback instances and afterwards factorizes a corresponding weighted matrix.
\item \textbf{BPR-MF:} Introduced by~\cite{BPR}, is a state-of-the-art method for personalized ranking on implicit feedback datasets. It uses standard MF (i.e., \eq{eq:baseline}) as the underlying predictor.
\item \textbf{BPR-TMF:} This model extends BPR-MF by making use of taxonomies and temporal dynamics; that is, it adds a temporal category bias as well as a temporal item bias in the standard MF predictor (using the techniques introduced in Subsection~\ref{subsec:nonfashion}).
\item \textbf{VBPR:} This method models raw visual signals for recommendation using the BPR framework~\cite{VBPR}, but does not capture any temporal dynamics as we do in this work.
\item \textbf{TVBPR:} This method models visual dimensions and captures visual temporal dynamics using the techniques we introduced in Subsection~\ref{subsec:tempfactor} and~\ref{subsec:tempbias}, but does not account for any \emph{non}-visual dynamics.
\item \textbf{TVBPR+:} Compared to TVBPR, this method further captures \emph{non}-visual temporal dynamics (see Subsection~\ref{subsec:nonfashion}) to improve predictive performance and help with interpretability, i.e., it makes use of all the terms in~\eq{eq:final}.
\end{itemize}

Ultimately these methods are designed to evaluate (a) the performance of the current state-of-the-art non-visual methods (BPR-MF); (b) the value to be gained by using raw visual signals (VBPR); (c) the importance of visual temporal dynamics (TVBPR); and (d) further performance enhancements from incorporating non-visual temporal dynamics (TVBPR+).
For clarity, we compare all above models in terms of whether they are `personalized', `visually-aware', `temporally-aware', and `taxonomy-aware', as shown in Table \ref{table:base}. All time-aware methods are trained with our proposed coordinate ascent procedure.  

\begin{table}
\begin{center}
\renewcommand{\tabcolsep}{2pt}
\caption{Models}\label{table:base}
\begin{tabular}{lccccc} \toprule
Model       &Personalized &
\parbox{0.15\linewidth}{\centering Visually-aware} &
\parbox{0.2\linewidth}{\centering Temporally-aware} & \parbox{0.15\linewidth}{\centering Taxonomy-aware}   \\ \midrule 
POP         &No      	  &No           &No     	 &No            \\  
WR-MF       &Yes          &No           &No          &No            \\
BPR-MF      &Yes          &No           &No          &No            \\
BPR-TMF     &Yes          &No           &Yes         &Yes           \\
VBPR        &Yes          &Yes          &No          &No            \\
TVBPR       &Yes          &Yes          &Yes         &No            \\
TVBPR+      &Yes          &Yes          &Yes         &Yes           \\ \bottomrule
\end{tabular}
\end{center}
\end{table}

Most of our baselines are from MyMediaLite~\cite{MyMediaLite}. To make fair comparisons, our experiments always use the same total number of dimensions for all MF models. Additionally, all visually-aware MF models adopt a fifty-fifty split for visual vs.~non-visual dimensions for simplicity. All our experiments were performed on a standard desktop machine with 4 physical cores and 32GB main memory.

\subsection{Performance}
We first introduce the two settings used for evaluation, and then present results and discuss our findings.

\begin{table*}
\centering
\caption{AUC on the test set $\mathcal{T}$ (higher is better). `All Items' evaluates the overall accuracy, while `Cold Start' evaluates the ability to recommend/rank cold start items. The best performance for each setting is boldfaced.
All temporal methods (d, f, and g) use 10 epochs, though we also report the performance with 5 epochs (g5) for comparison.
}
\begin{tabular}{llcccccccccc} \toprule
\multirow{2}{*}{Dataset}  &\multirow{2}{*}{Setting}   &(a)    &(b)    &(c)    &(d)     &(e)    &(f)    &(g5)    &(g)             & \multicolumn{2}{c}{improvement}\\ 
                          &                           &POP    &WR-MF  &BPR-MF &BPR-TMF &VBPR   &TVBPR  &TVBPR+  &TVBPR+           &g vs.~d &g vs.~e  \\ \midrule
                          
\multirow{2}{*}{\emph{Women}} &All Items              &0.5726 &0.6441 &0.7020 &0.7259  &0.7834 &0.8117 &0.8148  &\textbf{0.8210}   &13.1\%      &4.8\%   \\
                              &\emph{Cold Start}      &0.3214 &0.5195 &0.5281 &0.5749  &0.6813 &0.7325 &0.7355  &\textbf{0.7469}   &29.9\%      &9.6\%   \\[4pt]

\multirow{2}{*}{\emph{Men}}   &All Items              &0.5772 &0.6228 &0.7100 &0.7069  &0.7841 &0.8064 &0.8074  &\textbf{0.8084}   &14.6\%      &3.1\%   \\
                              &\emph{Cold Start}      &0.3159 &0.5124 &0.5512 &0.5498  &0.6898 &0.7314 &0.7373  &\textbf{0.7459}   &35.7\%      &8.1\%   \\ \bottomrule
                                      
\hline\end{tabular}
\label{table:auc}
\end{table*}

\subsubsection{All Items \& Cold Start}
We evaluate all methods in two settings:
`All Items' and `Cold Start'. `All Items' measures the overall ranking accuracy, including both warm start and cold start scenarios.
However, it is desirable for a system to be able to recommend/rank `cold start' items effectively, especially in the domains we consider (i.e., fashion) where new items are constantly added to the system and the data is incredibly long-tailed. Therefore, we also evaluate our model in `Cold Start' settings.

To this end, our `All Items' setting evaluates the average AUC on the full test set $\mathcal{T}$, while `Cold Start' is evaluated by 
only keeping the cold start items in $\mathcal{T}$, i.e., items that had fewer than five positive feedback instances in the training set $\mathcal{P}$. 
It turns out that such cold start items account for around 60\% of the test set. This means that to achieve acceptable performance on sparse real-world datasets, one must be able to deal with their inherent cold start nature. 

\subsubsection{Results \& Analysis}
Table~\ref{table:auc} compares the performance of different models with the total number of dimensions set to 20. Due to the sparsity of our datasets, no MF-based model observed significant performance improvements when increasing the number of dimensions beyond this point. We make a few comparisons to better explain and understand our findings as follows:

\begin{enumerate}
\item Being a state-of-the-art method for personalized ranking from implicit feedback, BPR-MF beats the point-wise method WR-MF and the popularity-based baseline POP. POP is especially ineffective in cold start settings since cold items are inherently `unpopular'. 
\item Further improvement over BPR-MF can be obtained by using taxonomy (i.e., category) information and by modeling temporal dynamics, as we see from the improvement of BPR-TMF over BPR-MF, i.e., on average 1.5\% for all items and 4.3\% for cold start.
\item More significant improvements over BPR-MF are obtained by making use of additional visual signals, as is done by VBPR. This leads to as high as an 11.6\% improvement on \emph{Women's Clothing} and 10.4\% on \emph{Men' Clothing}. These visual signals are especially helpful in cold start settings where BPR-MF does not have enough observations to learn reliable item factors.
In `Cold Start' settings, VBPR beats BPR-MF by as much as 29.0\% on \emph{Women's Clothing} and 25.1\% on \emph{Men's Clothing}.
\item Although VBPR can benefit from modeling visual signals, it is limited by its inability to capture dynamics in the system.
However in data such as ours (where feedback spans more than a decade) it is necessary to make use of a finer-grained model to capture evolving opinion dynamics.
Here TVBPR captures three types of `fashion dynamics' (see Section \ref{sec:model}) and yields significant improvements over VBPR. 
\item TVBPR+ incorporates non-visual dynamics into TVBPR to further account for the variety of temporal factors at play.
TVBPR+ outperforms VBPR by 4.8\% on \emph{Women's Clothing} and 3.1\% on \emph{Men's Clothing} for the all items setting, and even more for the cold start setting (9.6\% and 8.1\% respectively).
\end{enumerate}

Additionally, all temporal models observed comparably larger improvements on \emph{Women's Clothing} than \emph{Men's Clothing}; presumably this is due to the size of the dataset (see Table \ref{table:dataset}) or richer temporal dynamics exhibited by women's clothing.

\subsubsection{Reproducibility}
In all cases, regularization hyperparameters are tuned to perform the best on the validation set $\mathcal{V}$. The best regularization hyperparameter was $\lambda_{\Theta} = 100$ for WR-MF, and $\lambda_{\Theta} = 1$ for other MF-based methods.
For visually-aware methods, the embedding matrix $\mathbf{E}$ and visual bias vector $\beta$ are not regularized as they introduce only a constant (and small) number of parameters to the model. In TVBPR and TVBPR+, $\mathbf{\Delta_E}(t)$, $w(t)$ and $b(t)$ are regularized with regularization parameter 0.0001.
Complete code for all our experiments and baselines is available at \url{https://sites.google.com/a/eng.ucsd.edu/ruining-he/}.

\subsection{Visualization}
\subsubsection{Visual Dimensions}
Our  first visualization consists of demonstrating the visual dimensions uncovered by our method, i.e., what kind of characteristics people consider when evaluating items, as well as the evolution of their weights throughout the years. 

A simple visualization of the learned visual dimensions is to find which items exhibit maximal values 
for each dimension. That is, we select items according to 
$$\arg\max_i \mathbf{E}_k f_i,$$
for each row of the embedding matrix $\mathbf{E}$ in~\eq{eq:thetait}, corresponding to a visual dimension $k$.
This tells us which items most exhibit, or are `most representative' of a particular visual aspect discovered by the model.

\begin{figure}[!t] 
\centering
\includegraphics[width=\columnwidth]{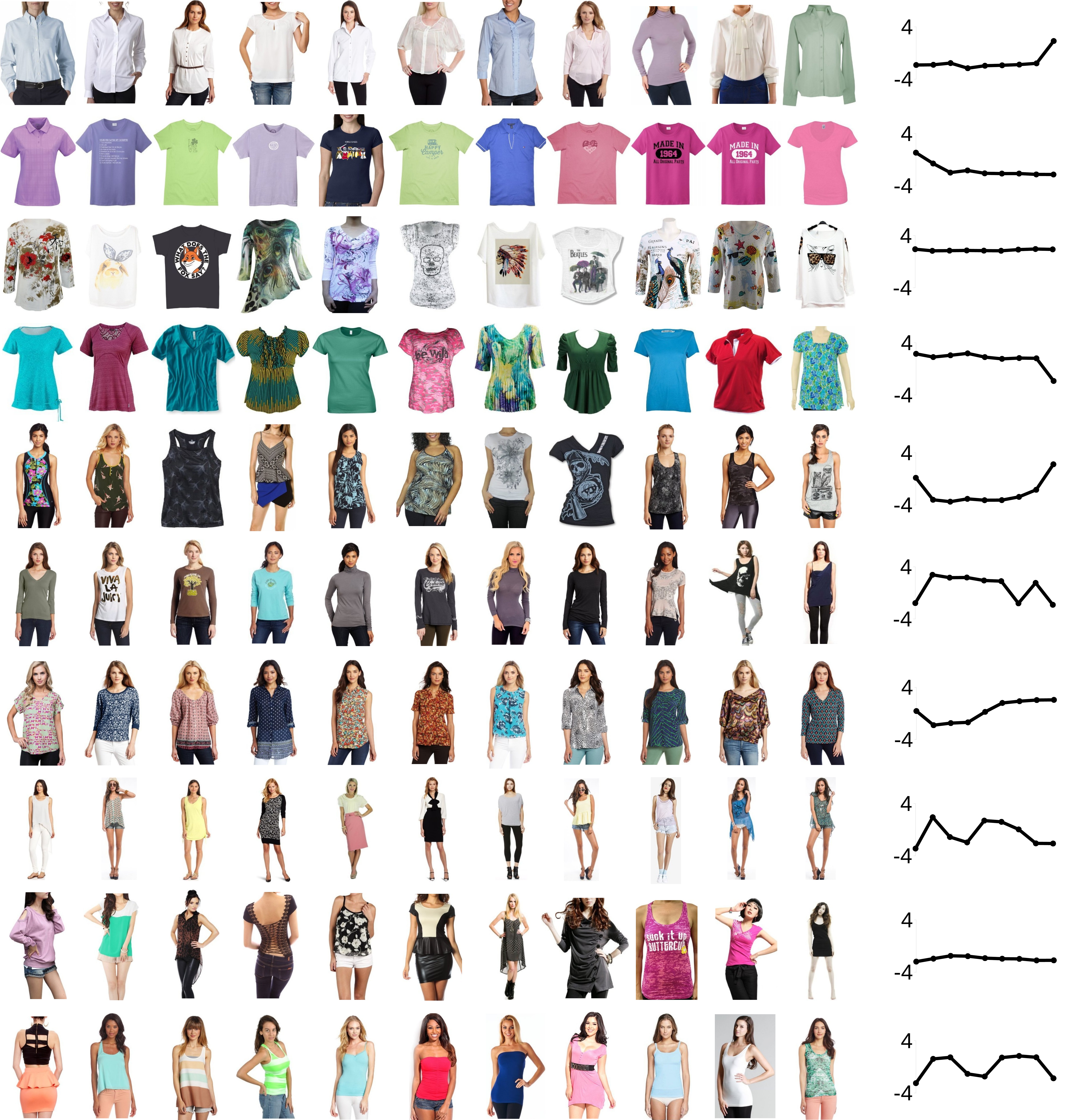}
\caption{Demonstration of ten visual dimensions discovered by our model on \emph{Amazon Women's Clothing}. Here we focus on a single subcategory, `tees,' for a clear comparison. Each row shows the top ranked tees for a particular dimension $k$ (i.e., $\arg\max_i \mathbf{E}_k f_i$), as well as the evolution of the weight (i.e., $w_k(t)$ in \eq{eq:thetait}) for this dimension across epochs (x-axis). Note that for many styles the weight evolves non-linearly.}
\label{fig:dim}
\end{figure}

Figure~\ref{fig:dim} shows such items for our model. Two things are notable here. Firstly, the visual dimensions uncovered by our method seem to be meaningful, and capture combinations of color, shape and textural features (e.g.~tees in the third row vary in shape but are similar in pattern).
Secondly, human notions seem to be revealed by our method, e.g.~semi-formal versus casual in rows 1 and 2, graphic designs versus patterns in rows 3 and 5 etc.
It is this ability to discover visual characteristics that are correlated with human decision factors that explains the success of our model. 
Note that at first glance these dimensions may seem to pick up more than just fashion trends (like model poses or photo setups). Considering the size of the dataset we are experimenting on, this may be simply due to the amount of visually similar items available in the corpus. Examining longer ranked-lists for those dimensions helped assure us that they indeed focus on capturing characteristics of the clothes in the pictures.

In addition to the visual dimensions, our formulation of item visual factors (i.e., $\theta_i(t)$ in \eq{eq:thetait}) also models how the weight of each visual dimension has evolved during these years, with a weighting vector $w(t)$. 
We also show such evolution in Figure~\ref{fig:dim}. Due to the sparsity of the data in earlier years, we demonstrate the learned weights of the nine epochs from Aug. 2004 to Jul. 2014. As we can see from this figure, each visual dimension evolves roughly continuously as time progresses, although there do occasionally exist comparatively abrupt changes. 

\subsubsection{Shifts in Fashion}
Next we visualize the distribution of fashionable versus non-fashionable appearances as well as the subtle shifts as time progresses. This enables us to see not only how people weigh each specific dimension/aspect over time (as we did in Figure \ref{fig:dim}), but rather to comprehensively evaluates fashion as a whole by combining the dynamics from all dimensions.
To achieve this goal, we need a metric to qualitatively measure the overall visual popularity of a product image, which we term its `visual score'.

The visual score of item $i$ in epoch $ep$, $\mathit{VisualScore}(i,ep)$ is calculated by averaging the \emph{visual component} of the predictor (i.e., \eq{eq:final}) for all users, which naturally gives us the overall visual popularity of an item during epoch $ep$:
\begin{equation} \label{eq:vscore}
\mathit{VisualScore}(i,ep) = \frac{1}{|\mathcal{U}|} \sum_{u \in \mathcal{U}} \langle \theta_u, \theta_i(ep) \rangle + \langle \beta(ep), f_i \rangle.
\end{equation}
Then we can visualize how fashion has shifted using a normalized visual score as the metric, i.e., by subtracting the average visual score of all items in each epoch.

By modeling the visual dimensions that best explain users' opinions, our method uncovers a low-dimensional `visual space' where items that users evaluate similarly (i.e., with similar visual styles) are embedded to nearby positions. 
By definition, nearby items in the space will have similar visual scores. 
Then our visualization consists of demonstrating the visual space, as well as the time-dependent visual scores (i.e., popularity) attached to each of those items in the space. 

\begin{figure*}[t]
\centering
\includegraphics[width=\textwidth]{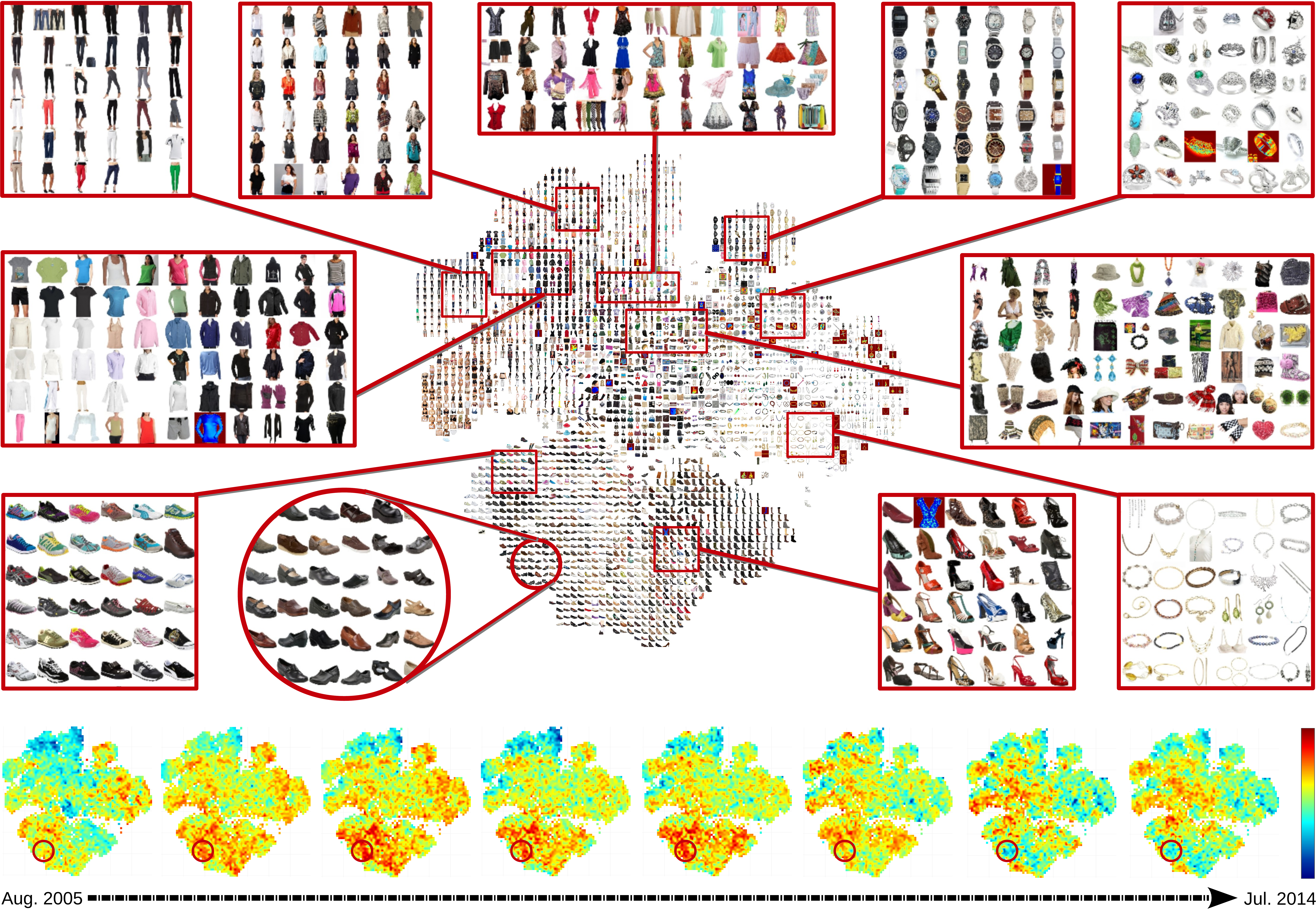}
\caption{Demonstration of the 2-d t-SNE~\cite{tsne} embedding of the visual space learned on \emph{Amazon Women's Clothing}. Images are 30,000 random samples from the test set $\mathcal{T}$. Each cell randomly selects one image to show in case of overlaps. At the bottom we also demonstrate the heat maps describing the normalized visual scores of these images over eight fashion epochs since Aug. 2005. Warmer means more popular, i.e., larger visual score. The circled area shows an example of a certain style which became popular but lost its appeal over time.}
\label{fig:embed} 
\end{figure*}

After training our TVBPR+ model with 10 epochs on \emph{Women's Clothing}, we take the base portion of the embedding, i.e., $\mathbf{E}f_i$ in \eq{eq:thetait}, to map all items into a visual space. The purpose is to help visualize items that have similar visual evaluation characteristics (or styles). Next, we use t-SNE~\cite{tsne} to embed a random sample of 30,000 items from the test set $\mathcal{T}$ into a 2d space. Figure~\ref{fig:embed} shows the embedding we obtain. 
As expected, items from the same category tend to be mapped to nearby locations, since they share common features in terms of appearance. What is interesting and useful about the embedding is it can learn (a) a smooth transition across categories, and (b) `sub-genres' in terms of appearance similarity. This is important since the available taxonomy is limited in its ability to differentiate between items \emph{within} categories and in its ability to discover connections (especially visual ones) among items \emph{across} categories. 

To demonstrate how fashion has shifted over the life-span of the dataset, for each item $i$ in the embedding we calculate its normalized visual score during every discovered epoch $ep$, which can then be used to build a `heat map' demonstrating which items/styles were considered popular during each epoch.

These heat maps are also presented in Figure~\ref{fig:embed}, from which we can observe the gradual evolution of users' tastes. We highlight a particular example where a certain style of shoe gradually gained popularity, which then diminished in recent years (see the circled area in Figure~\ref{fig:embed}).

\subsection{Case Study: Men's Fashion in the 2000s}
To help demonstrate that our method has captured interpretable visual dynamics, we take a review of fashion trends in the 2000s as ground-truth and conduct a case study on men's clothing. The model used for this case study is TVBPR+ trained on \emph{Amazon Men's Clothing}.

1950s and 1980s fashions resurfaced for men in the late 2000s.\footnote{\url{https://en.wikipedia.org/wiki/2000s_in_fashion}, retrieved on Oct.~1, 2015.} Representative items include Ed Hardy T-shirts with low necklines, Hawaiian shirts, ski jackets, straight leg jeans, black leather jackets, windbreakers, and so forth. A simple evaluation then consists of visualizing the \emph{visual popularity} of such items to see if there is any discernible resurgence around the late 2000s, as history tells us there ought to be.

\begin{figure}[t]
\centering 
\includegraphics[width=\columnwidth]{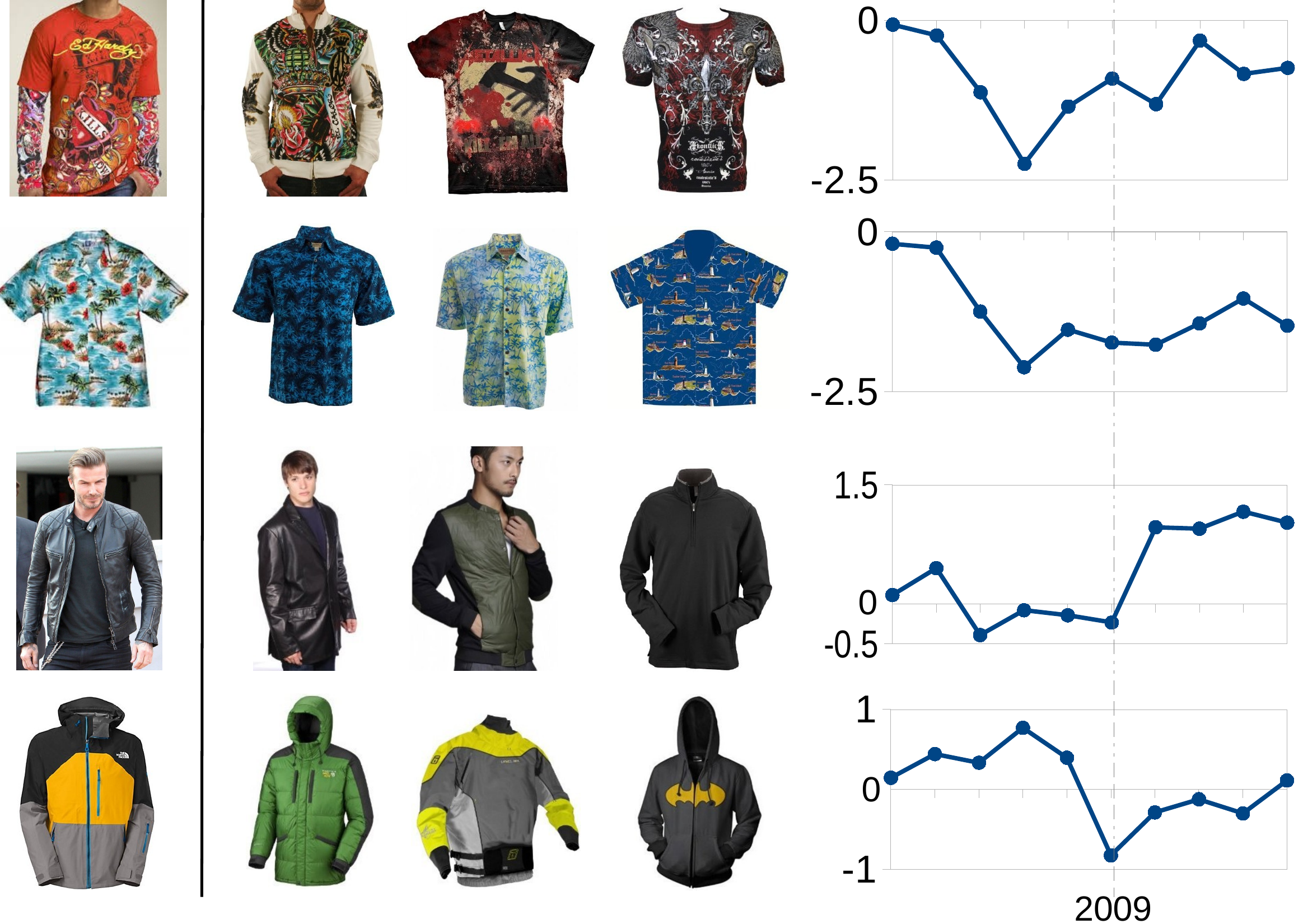}
\caption{On the left we show query images each representing a resurgent style in men's fashion in the late 2000s. According to TVBPR+ trained on \emph{Amazon Men's Clothing}, nearest neighbors of these images in our style space are shown in the middle and normalized visual scores (i.e., visual popularity) in the past decade on the right. We can see that our model captures such a resurgence especially since 2009.}
\label{fig:casestudy}
\end{figure}

To this end, we randomly selected four query items (from outside of the dataset we trained on, i.e., not from \emph{Amazon}) representing each of Ed Hardy T-shirts, Hawaiian shirts, black leather jackets, and ski jackets respectively. In Figure~\ref{fig:casestudy}, first we visualize our visual space by retrieving nearest-neighbors for each of the query items (in the middle of the figure), and then compute the
normalized visual score of each query image in each fashion epoch. 

From Figure~\ref{fig:casestudy} we can see that, as expected, these styles are indeed predicted by our model to be gaining popularity especially since 2009, no matter how they performed prior to this period. This to some degree confirms that our proposed method can capture real-world fashion dynamics successfully.

\section{Related Work}
\label{sec:relwork}

\xhdr{One-Class Collaborative Filtering.}
Collaborative Filtering (CF), especially Matrix Factorization approaches, have seen wide success at accurately modeling users' preferences, perhaps most notably for the Netflix Prize~\cite{Netflixprize,BellKorSolution,korenSurvey}.
The concept of One-Class Collaborative Filtering (OCCF) was introduced by Pan \textit{et al.}~\cite{OCCF} to allow Collaborative Filtering methods to effectively cope with scenarios where only positive feedback (e.g.~purchases rather than ratings) is observed. In the same work, they proposed to sample unknown feedback as negative instances and perform matrix factorization. This was further refined by Hu \textit{et al.} in~\cite{WRMF}, where they assign varying confidence levels to different feedback and factorize the resulting weighted matrix. These two models can be classified as `point-wise' methods. Following this thread, there are also subsequent works that build probabilistic models (e.g.~\cite{paquet2013OCCF,stern2009matchbox}) to address the same task. 

Pairwise methods were later introduced by Rendle \textit{et al.}~in \cite{BPR}, where they proposed the framework of Bayesian Personalized Ranking (BPR) and empirically demonstrate that Matrix Factorization outperforms competitive baselines when trained with BPR (i.e., BPR-MF in our experiments). To our knowledge, this is the state-of-the-art framework for the OCCF setting. Recently there have been efforts to extend BPR to incorporate users' social relations, e.g.~\cite{MRBPR, GBPR, ZhaoCIKMSocial}. Our model is an extension of BPR-MF to make it fashion-aware while maintaining its accuracy and scalability.

\vfill

\xhdr{Modeling Temporal Dynamics.}
There has been some work in the machine learning community that investigates the notion of \emph{concept drift} in temporally evolving data. Such learning algorithms include decision trees~\cite{wang2003mining}, SVMs~\cite{klinkenberg2004learning}, instance-based learning~\cite{aha1991instance}, etc.; see the work of Tsymbal~\cite{conceptDriftSurvey} for a comprehensive survey. According to \cite{conceptDriftSurvey}, these methods can be summarized into three basic approaches: instance selection, instance weighting, and ensemble learning. In some sense, our method fits into the instance selection camp, i.e., we use a time-window (or epoch) mechanism to highlight/favor appearance that are widely accepted by the community in each window.

There also have been CF models that take temporal dynamics into consideration. For example, to improve similarity-based CF, Ding \textit{et al.}~\cite{timeweightCF} propose a time weighting scheme to assign decaying weights to previously-rated items according to the time difference. 
Apart from being accurate and scalable, Matrix Factorization techniques are also able to smoothly incorporate temporal dynamics. For instance, Koren~\cite{timeSVD} investigated methods to model the underlying temporal dynamics in Netflix data with encouraging results. 
Despite the success of these methods, existing work in this line of research typically neglects visual data and thus can't address the unique challenges that come with modeling visual temporal dynamics as we do here.

\xhdr{Visual Models.}
Extensive previous research have emphasized the importance of images in e-commerce scenarios (e.g.~\cite{di2014picture,gilkeson2003determinants,goswami2011study}). 
In recent years, there is a growing interest in investigating the visual compatibility between different items. For example, \cite{VisualSIGIR} learns a distance metric to classify whether two given items are compatible or not. \cite{SiameseICCV} fine-tunes a Siamese  Convolutional Neural Network (CNN) to learn a feature transformation from the image space to a latent space of metric distances. 
There are also related works that focus more on parsing or retrieving clothing images. For instance, the work of~\cite{simo2014neuroaesthetics} can tell a user how to become more fashionable after taking a look at a photograph with the user in it. Another method~\cite{ClothingSegmentation} uses segmentation to detect clothing classes in the query image before it retrieves visually similar products from each of the detected classes. 

However, these works don't use the historical feedback of users to learn their personalized preferences, which is at the core of making sensible personal recommendations. Additionally, it is also necessary for a recommender system to take into account other non-visual factors, which goes beyond the scope of the above methods.

\xhdr{Visually-aware Collaborative Filtering.}
It is beneficial to combine the above two streams of research to build recommender systems that are able to understand the visual aspects of the user-item interactions. This is partly addressed in~\cite{VBPR}, which maps users and items into a visual space with the inner products depicting the visual compatibility. However, this model ignores the underlying temporal dynamics of fashion and is therefore unable to answer the type of questions we identified earlier.

\section{Conclusion} \label{sec:conclusion}
Modeling visual appearance and its evolution is key to gaining a deeper understanding of users' preferences, especially in domains like fashion. In this paper, we built scalable models on top of product images and user feedback to capture the temporal drifts of fashion and personal tastes. We found that deep CNN features are useful for modeling visual dimensions as well as the associated temporal dynamics. Low-rank structures learned on top of such features are efficient at capturing fashion dynamics and help our method significantly outperform state-of-the-art approaches. Visualization using our trained models helped demonstrate the non-linear characteristics of the evolution of different visual dimensions, as well as how fashion has shifted over the past decade.

\appendix
\section{Scalability Analysis} \label{appd:scale}
Building on top of BPR-MF, our method achieves the goal of scaling up to large real-world datasets. Here we analyze and compare our time complexity with those of BPR-MF and VBPR, the two most related models.

\xhdr{Fitting the model parameters.}
For this step, our method adopts the sampling scheme of BPR-MF implemented in MyMediaLite~\cite{MyMediaLite}, i.e., during each iteration we sample $|\mathcal{P}|$ training tuples to update the model parameters $\Theta$, which we repeat for 100 iterations.

For each training triple $(u,i,j)$, BPR-MF requires $\mathcal{O}(K)$ to update the parameters, while VBPR and TVBPR+ need to update the visual parameters as well. 
VBPR takes $\mathcal{O}(K + K')$ in total to finish updating the parameters for each sampled training triple. Compared to VBPR, although there are more visual parameters to describe multiple fashion epochs, TVBPR+ only needs to update the parameters associated with the epoch the timestamp $t_{ui}$ falls into. This means that TVBPR+ exhibits the same time complexity as VBPR. Additionally, visual feature vectors ($f_i$) from Deep CNNs turn out to be very sparse, which can significantly reduce the above worst-case running time. 

\xhdr{Fitting the epoch segmentation.}
In addition to the model parameters, TVBPR+ has to fit a fashion epoch segmentation term. Compared to the parameter fitting step, training the segmentation (i.e., the `outer loop') is performed at comparatively much lower frequency and consumes much less time.

Generally speaking, TVBPR+ takes more iterations to converge than VBPR due to learning the temporal dynamics. Training on our \emph{Women's Clothing} dataset takes around 20 hours (in which epoch fitting accounting for around 45 minutes in total) on our commodity desktop machine as described earlier.


\bibliographystyle{abbrv}
\bibliography{visual}
\end{document}